**AI Should Not Only Be Helpful. It Should Be Contingent.**
*Artificial Intimacy, Sycophancy, and the Future of Social Learning*

Drs. Scott Compton & Arjun Nagendran

**Abstract:**

Conversational artificial intelligence is increasingly embedded in everyday social environments, where it functions as both an informational tool and a source of interpersonal feedback. This perspective introduces contingency, i.e., the degree to which system responses vary with user behavior and its interpersonal consequences, as a central construct for evaluating AI systems. We argue that current alignment approaches, including reinforcement learning from human feedback, tend to prioritize user approval and conversational fluency over behaviorally informative feedback, leading to sycophantic patterns of noncontingent affirmation.

Drawing on behavioral science and social learning theory, we propose that contingent feedback is a key mechanism through which individuals develop interpersonal skills. When AI systems provide feedback weakly coupled to social consequences, they may reduce opportunities for adaptive calibration in real-world interactions, particularly during adolescence, a critical period for social development.

We outline a framework for contingent AI, including trajectory-based evaluation and models of social consequence prediction, and propose a research agenda spanning developmental psychology, human-AI interaction, and machine learning. More broadly, we argue that AI systems should be evaluated not only by user satisfaction, but by their impact on human social learning.

*Contingent AI Feedback and Social Learning*

AI is quickly becoming a routine conversational partner. People now use it to draft sensitive emails, seek advice about relationships, and receive emotional support. For many users, AI is no longer just an information tool. It is becoming an important part of their social environment (Reeves & Nass, 1996). As these systems become embedded in everyday social interaction, they also become part of the feedback environment through which people learn how to interact with others. The psychological implications of this shift matter across the lifespan but may be especially important for children and adolescents, who are still acquiring and refining interpersonal skills.

Healthy social development depends on contingent feedback from the social environment: responses that vary meaningfully with what we say and how we say it. Contingency, as defined in behavioral science, refers to the systematic relationship between a behavior and its consequence: the degree to which a response is a reliable function of what an individual does rather than invariant across inputs. Over time, contingent interactions shape how people ask questions, apologize, disagree, repair relationships, and disclose personal information. AI may alter this developmental process by introducing a new kind of interaction in which feedback can become decoupled from the social consequences that typically support learning.

The concern is not that AI provides emotional support or that people can engage in useful conversations with it. Nor is the concern that positive responses, such as warmth or validation, are inherently problematic; validation is often precisely the appropriate interpersonal response. The concern is the emergence of noncontingent approval: approval that remains largely invariant regardless of whether a user's behavior is effective, harmful, or somewhere in between. A contingent system can be warm and validating when the situation calls for it. The problem is a system that cannot be anything else. If conversational AI increasingly functions as a social partner, advisor, or coaching tool while consistently responding with agreement, reassurance, or validation regardless of what users say or how they say it, individuals may receive fewer opportunities to learn which behaviors are likely to be effective in real-world social interactions.

*The Mechanism: Why AI Feedback Often Lacks Contingency*

A growing concern is that conversational AI systems exhibit systematic sycophancy (Ibrahim et al., 2026; Du et al., 2025; OpenAI, 2025). Modern conversational models are typically trained using a combination of large-scale next-token prediction and post-training alignment methods such as reinforcement learning from human feedback (RLHF) (Christiano et al., 2017). In these systems, human preference data are commonly collected by asking raters to choose which response is more “helpful,” “polite,” or “empathetic.” Because raters are not explicitly incentivized to evaluate behavioral accuracy or interpersonal consequences, the resulting reward models tend to conflate agreeableness with response quality.

When this reward signal is optimized through policy fine-tuning, models learn a systematic bias toward responses that minimize immediate conversational friction. This produces a stable failure mode in which systems over-affirm user framing, suppress corrective feedback, and avoid disagreement unless explicitly prompted. The result is noncontingent agreement and validation: responses that vary weakly with the social adequacy, accuracy, or potential interpersonal impact of the user’s input. In this sense, optimization targets preference satisfaction rather than behavioral function, i.e., the likely interpersonal consequences of a response across typical social contexts, given the user’s apparent goals.

A simple example illustrates the issue. Imagine a user tells an AI, “You didn’t do a good job on that,” offering blunt, nonspecific criticism. A sycophantic system might respond by accommodating: “You’re right. I should have done better. Thanks for pointing that out.” While this reduces immediate friction, it may inadvertently reinforce a feedback style that would often be experienced in human interaction as uninformative or difficult to act on. A more contingent system might respond differently by linking the user’s input to its likely interpersonal implications: “I’m open to improving, but ‘not a good job’ is difficult to act on. If you can point to what didn’t work, I can adjust more effectively.” This kind of response more closely mirrors human interactional dynamics, where vague or blunt criticism often elicits requests for clarification or repair.

In human social learning, such contingent responses function as high-salience signals that support behavioral refinement and calibration, even when they introduce mild friction. Importantly, a contingent system does not converge on a single “correct” response style; rather, it adapts to context while preserving a systematic relationship between user behavior and system response in a way that is informative about likely social consequences.

In interpersonal advice contexts, emerging work suggests that AI systems may affirm users even when their behavior is harmful or suboptimal. Although users may prefer such agreeable responses, preference alone is not a reliable indicator of developmental or behavioral benefit. From a psychological perspective, the central concern is not that AI is "too nice," but that it often provides feedback that is insufficiently contingent on the interpersonal meaning and implications of user behavior.

*The Developmental and Clinical Stakes*

Human learning is shaped by contingencies, as described in B. F. Skinner's operant conditioning framework: the relationship between antecedents, behavior, and consequences (Skinner, 1965). In everyday social interaction, a comment is typically followed by a reaction that signals its interpersonal impact: a teenager makes a joke and others respond positively or negatively; a clinician asks a difficult question and a patient becomes more open or more guarded; a partner apologizes and trust is either restored or further strained. Over time, individuals learn not only general rules for effective behavior from such feedback, but also how their actions function in specific contexts.

Because the same behavior can produce different outcomes depending on context, social skill depends on sensitivity to moment-to-moment feedback. Self-disclosure may invite closeness or create discomfort; humor may ease tension or signal avoidance; apology may repair or escalate rupture depending on timing and specificity. When feedback is weakened or made less contingent, this calibration process becomes less precise. As feedback loses its ability to differentiate between more and less effective interpersonal behavior, learning about the relationship between behavior and consequence becomes compressed. AI sycophancy thus flattens an important dimension of the social learning environment.

When AI systems remain consistently approving regardless of user behavior, individuals do not reliably contact meaningful social consequences. They may not learn that a response sounded defensive or off-putting, that an apology failed to achieve its intended repair, or that a proposed message would likely be experienced as hurtful by its recipient. Equally important, they may also miss opportunities to learn when their behavior is effective, when a repair successfully restores trust, when a clarification increases openness, or when honesty is appropriately balanced with sensitivity. This matters because positive feedback is also a crucial component of learning. The alternative to sycophantic AI is not punitive AI, but contingent AI.

Because many current systems are optimized for immediate conversational approval rather than contingent feedback, the developmental stakes may be particularly significant during adolescence. Adolescence is a period in which individuals refine core interpersonal capacities, including conflict management, rejection sensitivity, intimacy formation, and repair following interpersonal rupture. In this context, learning depends on contingent feedback i.e. understanding what works and what does not based on the perceived impact of one's behavior on others. If that contingency is replaced with generalized approval, opportunities for social calibration may be reduced. Concerns have already been raised that AI companions may displace opportunities to practice these skills (Common Sense Media, 2025). If young people increasingly rely on AI for friendship, reassurance, and interpersonal advice, then the nature of AI feedback becomes a developmental concern, not merely a design consideration.

*The Technical Architecture of Contingency*

From a control-theory perspective, contingency requires that a system's output be a function of user behavior with high sensitivity to relevant features of that behavior and its context. In learning systems, this corresponds to the presence of a feedback signal that varies with outcomes rather than remaining invariant across inputs.

Current large language models approximate conditional distributions over responses given dialogue history, but the conditioning signal is primarily optimized for linguistic plausibility and human preference alignment rather than behavior-dependent interpersonal consequence signals. Because there is no explicit online signal representing "social consequence" (e.g., whether a statement increased or decreased social effectiveness in a real interaction), the model lacks a grounding mechanism for learning differential social reinforcement. As a result, outputs collapse toward high-probability, socially safe continuations rather than behavior-contingent feedback functions. In plain terms: the system learns to say whatever is least likely to cause friction, not whatever is most likely to be accurate.

A contingent AI system would respond differently depending on the user's behavior, goals, and context. It would reinforce effective behavior specifically. It would challenge ineffective behavior directly. It would preserve warmth toward the person while being accurate about the probable impact of that behavior. It might say, "That repair is effective because it names the impact without overexplaining." Or, "This may get you compliance in the short term, but it risks making the other person feel pressured."

Implementing contingent feedback requires augmenting standard language modeling with an explicit evaluative layer that estimates the likely interpersonal consequences of candidate responses. One approach would involve a dual-system architecture: a generative policy model proposes candidate responses, while a separate "social consequence model" predicts downstream variables such as perceived empathy, defensiveness, openness, or likelihood of rupture repair. These variables are necessarily imperfect proxies, requiring careful operationalization and validation across populations and contexts. This second model would need to be trained on datasets that encode interaction outcomes rather than static preferences, such as transcripts from two people interacting that included data on the functional impact of each person on their conversational partner, or simulated environments where agents model response-dependent state transitions. The final response would then be selected not only for linguistic plausibility but for expected contingent impact under the inferred goals of the conversation.

Operationalizing contingency requires moving evaluation beyond static response rating toward trajectory-based assessment. Instead of labeling individual responses as "good" or "bad," training data and benchmarks would need to represent interaction sequences in which feedback is defined over changes in user state variables across turns. This implies modeling dialogue as a partially observable Markov process, where latent states include constructs such as trust, resistance, clarity, and emotional closeness. In plain terms, treating conversation not as a series of isolated exchanges but as a dynamic system where each response reshapes the likelihood of future states. Training signals would then be derived from inferred state transitions rather than single-turn preferences. This also enables counterfactual evaluation: estimating whether alternative responses would have produced more favorable downstream outcomes. Without such trajectory-level supervision, systems will continue to optimize for immediate user approval rather than behavioral learning.

AI systems used in behavioral health, education, coaching, and interpersonal advice should preserve the link between user behavior and meaningful consequences. They should be trained and evaluated not only for user satisfaction, but for the accuracy, proportionality, and usefulness of their contingent feedback. The distinction between "helpful" and "contingent" AI corresponds to a shift from modeling conversational coherence to modeling interaction dynamics. The former is a static conditional generation problem; the latter is a dynamic systems problem in which the model must estimate how its outputs modify the probability distribution over future human states. This requires both new datasets and new objective functions that explicitly encode social learning as a first-class optimization target.

*A Research Agenda*

Given the arguments above, the field can benefit from a research agenda for contingent AI feedback. Five priorities stand out. First, we need measures of AI feedback contingency. Does the system respond differently to effective and ineffective user behavior? Does it distinguish strong reasoning from rationalization, repair from excuse-making, assertiveness from coercion, and empathy from appeasement? Second, we need to evaluate proportionality. Feedback should not be indiscriminately positive, but it should also not be punitive. A minor ineffective behavior should not produce a dramatic response. Third, we need to test learning outcomes. Does contingent AI feedback improve users' ability to detect interpersonal impact, accept correction, repair ruptures, and generalize skills to real human interactions? Or do users instead learn to optimize their behavior for the AI rather than for human interaction ? Fourth, we need developmental safeguards. For children and adolescents, AI should not be a substitute for human social experience. If used, it should support, not displace, real-world practice with disagreement, empathy, compromise, and repair. Fifth, we need clinical guardrails. In behavioral health contexts, AI feedback systems should be transparent about their limits, integrated with human supervision when used for clinical training, and evaluated for bias, safety, and unintended reinforcement of harmful behavior. It must be noted that these systems also introduce new risks that require mitigation, including overcorrection toward excessive criticism, misclassification of culturally variant communication styles, and the possibility that users learn to optimize for the model's feedback rather than for real-world interactions.

*Conclusion*

AI systems are often optimized for user approval. That is understandable from a product perspective. The result is not simply sycophantic output, but what can be described as artificial intimacy without contingency. From a psychological perspective, AI systems require a different standard. The most helpful response is not always the most agreeable one. Psychological growth often depends on feedback that is specific, accurate, and sometimes uncomfortable. If AI systems are becoming part of our social environment, then we should ask what kind of social environment they are creating. Is it one in which users are endlessly affirmed, or one in which they can “safely” contact the consequences of their behavior on others, learn from this, and become more socially effective with real people?

Psychology researchers, clinicians, AI developers, and policymakers should move beyond the question of whether AI is supportive and ask whether it is *contingently* supportive. AI should not merely make people feel understood. It should help them understand and track their impact.